\pdfoutput=1

\documentclass[11pt]{article}

\usepackage[final]{acl}

\usepackage{times}
\usepackage{latexsym}

\usepackage[T1]{fontenc}

\usepackage[utf8]{inputenc}

\usepackage{microtype}

\usepackage{inconsolata}

\usepackage{graphicx}

\usepackage{xspace}
\usepackage{todonotes}
\usepackage{multirow}
\usepackage{algorithm}
\usepackage{algpseudocode}
\usepackage{hyperref}
\usepackage{url}
\usepackage{graphicx}
\usepackage{amsmath}
\usepackage{booktabs}
\usepackage{adjustbox}
\usepackage{enumitem}
\usepackage{colortbl}
\usepackage{xcolor}
\usepackage[most]{tcolorbox}
\usepackage{amsfonts}
\usepackage{subfig}
\usepackage{stfloats}

\newcommand{\one}{({\em i}\/)\xspace}
\newcommand{\two}{({\em ii}\/)\xspace}
\newcommand{\three}{({\em iii}\/)\xspace}

\def\eg{\emph{e.g.}\xspace}

\def\ie{\emph{i.e.}\xspace}

\def\model{MMKD\xspace}

\title{Multi-MLLM Knowledge Distillation for Out-of-Context News Detection}

\author{\textbf{Yimeng Gu$^{1}$, Zhao Tong$^{2}$, Ignacio Castro$^{1}$, Shu Wu$^{3}$, Gareth Tyson$^{1,4}$} \\
  $^1$Queen Mary University of London \\
  $^2$Institute of Information Engineering, Chinese Academy of Sciences \\
  $^3$Institute of Automation, Chinese Academy of Sciences \\
  $^4$The Hong Kong University of Science and Technology (GZ) \\
  \texttt{yimeng.gu@qmul.ac.uk}\\
  }

\begin{document}
\maketitle
\begin{abstract}
Multimodal out-of-context news is a type of misinformation in which the image is used outside of its original context. Many existing works have leveraged multimodal large language models (MLLMs) for detecting out-of-context news. However, observing the limited zero-shot performance of smaller MLLMs, they generally require label-rich fine-tuning and/or expensive API calls to GPT models to improve the performance, which is impractical in low-resource scenarios. In contrast, we aim to improve the performance of small MLLMs in a more label-efficient and cost-effective manner. To this end, we first prompt multiple teacher MLLMs to generate both label predictions and corresponding rationales, which collectively serve as the teachers' knowledge. We then introduce a two-stage knowledge distillation framework to transfer this knowledge to a student MLLM. In Stage 1, we apply LoRA fine-tuning to the student model using \emph{all} training data. In Stage 2, we further fine-tune the student model using both LoRA fine-tuning and DPO on the data points where teachers' predictions \emph{conflict}. This two-stage strategy reduces annotation costs and helps the student model uncover subtle patterns in more challenging cases. Experimental results demonstrate that our approach achieves state-of-the-art performance using less than 10\% labeled data.
\end{abstract}

\section{Introduction}

Multimodal out-of-context news is a common form of misinformation~\cite{fazio2020ooc} which re-purposes an image outside of its original context. For example, the caption``\emph{the safety of the newly released vaccine is still unclear}'' might be attached to an image of a Covid-19 vaccine, whilst the in-context image is actually about another vaccine. Such misinformation can mislead the public and even trigger social panic. Therefore, it is crucial to devise tools that can detect out-of-context news before it spreads. 

Multimodal large language models (MLLMs) are increasingly being used in out-of-context news detection. Some works use MLLMs to enhance explainability. \citet{qi2024sniffer} applies supervised fine-tuning on Instruct-BLIP to provide reasonable explanations. Similarly, \citet{shalabi2024leveraging} applies supervised fine-tuning on MiniGPT-4. Other works leverage proprietary MLLMs' reasoning capability in multi-agent debates~\citep{lakara2024mad}, and multi-agent collaboration~\citep{wu2025exclaim} to detect out-of-context news. 
These prior works~\citep{qi2024sniffer, shalabi2024leveraging, liu2024fka} recognize smaller MLLMs' limited performance when serving as a zero-shot out-of-context news detector. Therefore, previous approaches generally require either label-rich fine-tuning or expensive API call to enhance their performance, which makes them costly and less generalizable to low-budget scenarios, and thus unsuitable for real-world deployment.  
\begin{figure}[t]
  \centering
  \includegraphics[width=\columnwidth]{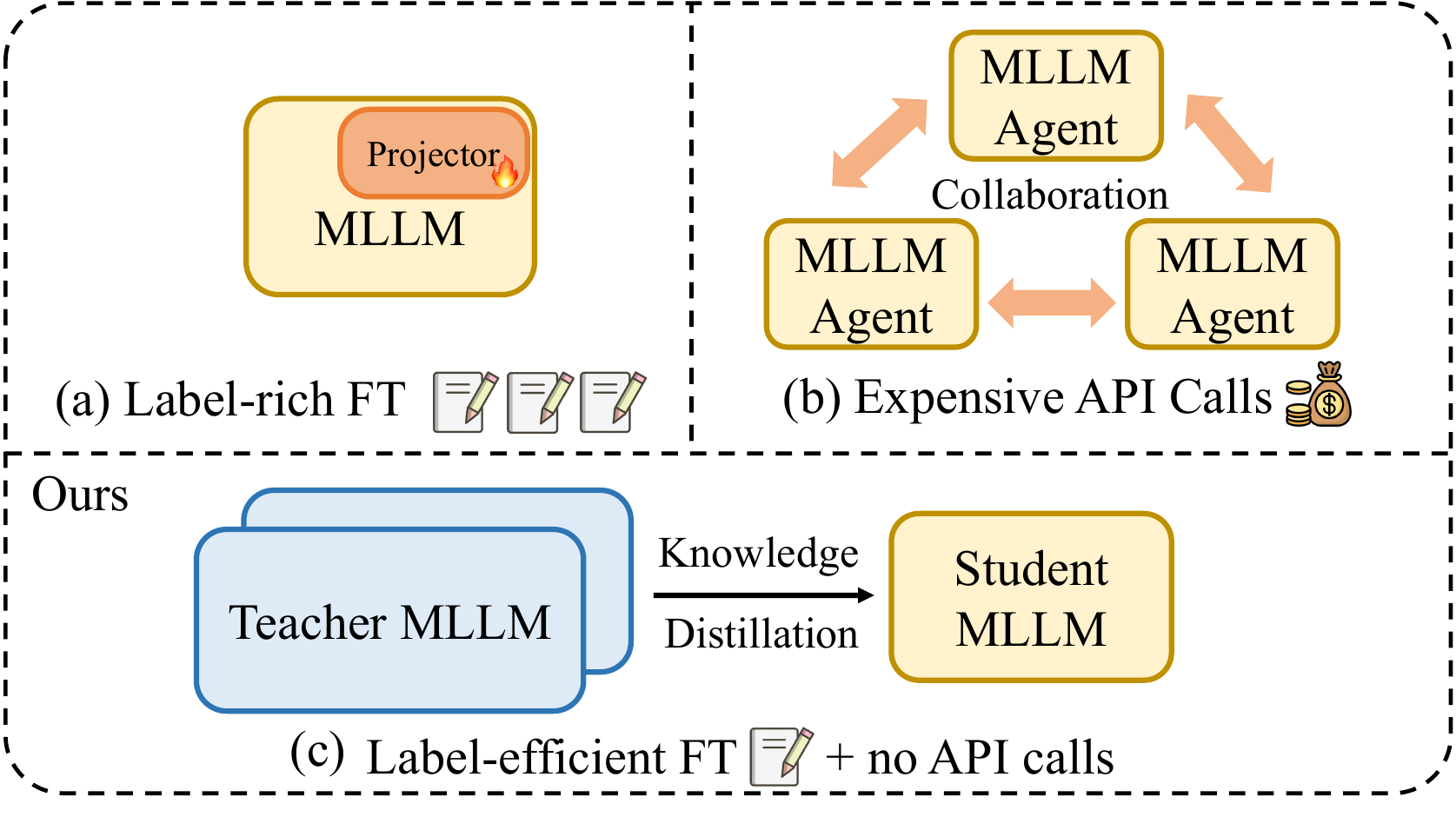}
  \caption{A brief comparison between our proposed approach and previous works. Previous approaches either (a) demands label-rich fine-tuning or (b) makes expensive API calls to proprietary MLLMs. In contrast, our approach only requires a few labels and does not use proprietary MLLMs.}
  \label{fig:overview}
\end{figure}

Contrary to prior works,
we aim to enhance small MLLM's performance in a more label-efficient and cost-effective manner (see Figure~\ref{fig:overview}).
Recently, several open-source models with \textasciitilde 70B parameters have exhibited superiority on the MMLU benchmarks~\citep{hendrycks2020measuring}. 
However, using such large models has significant computational cost, making them unsuitable in resource-limited environments.
As such, this motivates us to explore the potential of transferring knowledge to smaller MLLMs (\ie student models) which are more suitable for resource-limited settings. 

However, knowledge distillation on out-of-context news detection comes with two challenges. The \emph{first} challenge resides in effectively eliciting larger MLLMs' knowledge for out-of-context news detection. 
The reason is that, unlike traditional image retrieval tasks, news images and captions generally contain entities with more accurate semantics, \eg the specific place and event.
Some are crucial for matching the news image and caption, while others may cause an overemphasis on unimportant details. Hence, it is critical for the teacher model to establish appropriate judgment criteria and thus provide reliable knowledge.
\emph{Second}, it is challenging to distill the knowledge to the student model in a cost-efficient way. Note that student models are also MLLMs --- they typically contain billions of parameters. As a result, adjusting the entire model during knowledge distillation is computationally intensive, especially for platforms that build upon decentralized or low-resource infrastructure. Moreover, refining the model capability for this task involves training on labeled news image-caption pairs, and obtaining these annotations is costly. Considering the fast evolving nature of news, it is necessary to reduce the reliance on labeled data during knowledge distillation.

To address the above two challenges, we propose a novel multi-teacher knowledge distillation framework, \model, for out-of-context news detection. It first prompts multiple
teacher MLLMs to generate both label predictions and corresponding rationales. Inside the prompt, we incorporate web-retrieved evidence to help the MLLMs focus on the key elements relevant for consistency checking. This enriched context facilitates the elicitation of teacher MLLM's out-of-context news detection knowledge. 
Furthermore, to efficiently distill the knowledge to the student model, we fuse the knowledge of two teacher MLLMs using a two-stage knowledge distillation strategy. Specifically, we first apply Low-Rank Adaptation (LoRA) to fine-tune the student model on \emph{all} the training data, followed by both LoRA and Direct Preference Optimization (DPO) fine-tuning on the data points that teachers' opinions \emph{conflict}. Notably, label annotations are only required for the latter stage, which minimizes overheads. Through this two-stage fine-tuning process, the student model is able to effectively learn useful knowledge from both teachers.

To summarize, our main contributions are as follows.

\begin{itemize}

\item We propose a novel approach \model to improve the out-of-context news detection capability of a small student MLLM by distilling knowledge from multiple large teacher MLLMs, with only 1.2M learnable parameters.

\item \model incorporates evidence into the prompt to effectively elicit knowledge from teacher MLLMs, and employs a two-stage multi-teacher knowledge distillation strategy to fuse the knowledge from two teacher MLLMs.

\item The learned student model outperforms the state-of-the-art (SOTA) performance by 1.87\% in accuracy using just 10\% of the labeled data, while matching the teacher model's performance with 10$\times$ fewer parameters.

\end{itemize}

\section{Related Work}
\label{sec:related-work}

\subsection{Out-of-Context News Detection}
\label{sec:rw-ooc}

Prior approaches to out-of-context news detection can be divided into three categories: 
\one fine-tuning vision and language models (VLMs), \two incorporating auxiliary information, and \three leveraging the reasoning capability of MLLMs and LLMs.

Pioneering works involve fine-tuning VLMs. \citet{biamby-etal-2022-twitter} fine-tunes CLIP~\citep{radford2021learning} to detect out-of-context news. 
Similarly, \citet{luo-etal-2021-newsclippings} fine-tunes both CLIP and VisualBert~\citep{li2019visualbert}. \citet{abdelnabi2022open} additionally introduces the retrieved textual and visual evidences to assist the consistency check. 
Later works leverage auxiliary multimodal information. \citet{shalabi2023image} generates augmented images from text and augmented text from images. \citet{yuan-etal-2023-support} extracts stances from external multimodal evidence to enhance the detection. \citet{ma2024interpretable} uses neural symbolic model to enhance the model interpretability. Recent works have increasingly used LLMs and MLLMs. \citet{qi2024sniffer} and \citet{shalabi2024leveraging} respectively adopt two-stage instruction tuning on Instruct-BLIP~\cite{dai2024instructblip} and MiniGPT-4~\citep{zhu2023minigpt} to provide reasonable explanations alongside the prediction. \citet{lakara2024mad} employs GPT-4o~\citep{hurst2024gpt} for multi-agent debates.

Although prior works are effective in detecting out-of-context news, they are computationally intensive or expensive. In contrast, our work aims to minimize the computation cost by 
efficient fine-tuning and labeling.

\subsection{Knowledge Distillation for (M)LLMs}

The growing scale of LLMs has enabled them to achieve better performance. To compress the model size while maintaining its capability, many works~\cite{xu2024survey} explore knowledge distillation to transfer knowledge from a larger teacher LLM to a smaller student LLM. Leveraging KL-divergence, \citet{liu2024ddk} proposes a domain balance-aware knowledge distillation method for LLMs. \citet{wu-etal-2024-divide} teaches the student model how to decompose a question via fine-tuning (FT). \citet{shridhar-etal-2023-distilling}, \citet{li2023symbolic} and \citet{magister-etal-2023-teaching} transfer the step-by-step Chain-of-Thought (CoT) reasoning capabilities of larger models to a smaller model via FT. In addition to CoT, \citet{chenglin2024mixed} also distills knowledge through Program-of-Thought (PoT). In \citet{zhao-etal-2024-large-language}, a larger teacher model generates multiple answers and rationales, from which the correct ones are distilled into a smaller model through few-shot prompting. \citet{zhang2025distill} distills not only knowledge but also rewards to a student model via FT and RL respectively.

Our work is the first to investigate multi-teacher knowledge distillation in out-of-context news detection. In contrast to previous works, we innovatively apply DPO to fuse the multi-teacher knowledge.


\section{Task Definition}
\label{sec:task-definition}

Out-of-context news detection is the task of predicting whether a given news image-caption pair is out-of-context or in-context. Here, \emph{out-of-context} means that the news image comes from a different context than the news caption and is used outside of its original setting, while \emph{in-context} means that they are from the same context.

Formally, let $(x_{\mathrm{img}}, x_{\mathrm{cap}}) \in \mathcal{X}$ denote an image-caption pair, where $x_{\mathrm{img}}$ is the news image and $x_{\mathrm{cap}}$ is the news caption. The goal of out-of-context news detection is to predict its label $y \in \{0, 1\}$, where $y = 1$ indicates that the news image-caption pair is out-of-context, and $y = 0$ indicates that it is pristine.

\section{Methodology}

\begin{figure*}[htbp]
  \centering
  \includegraphics[width=\textwidth]{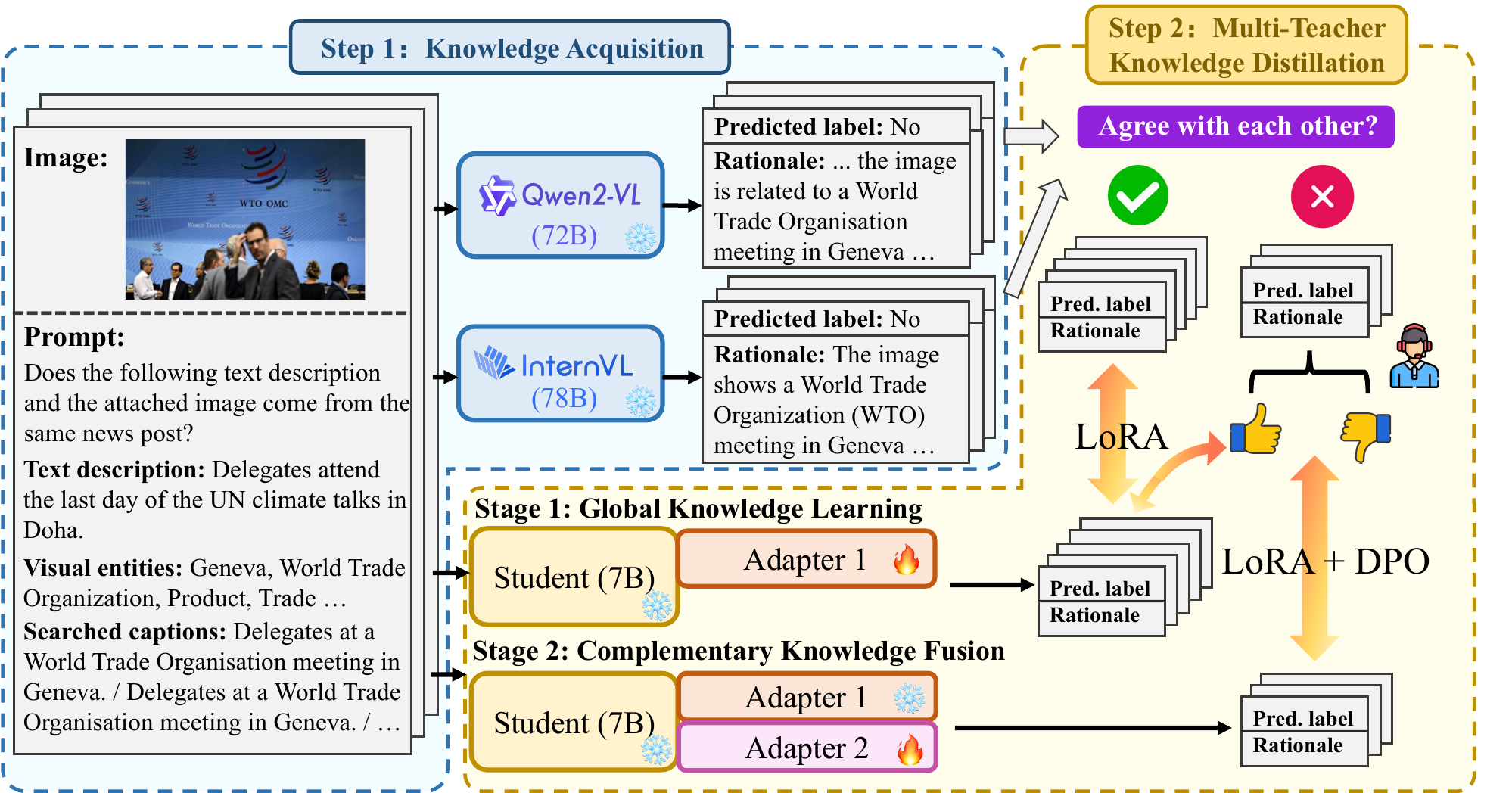}
  \caption{The framework of \model. It consists of two steps: \one \emph{Knowledge Acquisition}, which prompts teacher MLLMs to obtain predicted labels and corresponding rationales; and \two \emph{Multi-Teacher Knowledge Distillation}, which LoRA fine-tunes the student MLLM on the acquired teacher knowledge with two stages: Global Knowledge Learning and Complementary Knowledge Fusion.}
  \label{fig:model}
\end{figure*}

In this section, we present our proposed approach \model. The overall approach is illustrated in Figure~\ref{fig:model}. 
\model consists of two steps: \one \emph{Knowledge Acquisition}, which extracts predicted labels and corresponding rationales from teacher MLLMs; and \two \emph{Multi-Teacher Knowledge Distillation}, which LoRA fine-tunes the student MLLM on the acquired teacher knowledge. Specifically, the second step encompasses two stages: \emph{Global Knowledge Learning} and \emph{Complementary Knowledge Fusion}. Stage 1 equips the student model with reasonable logic to detect out-of-context news, while Stage 2 reinforces the student's capability on ``hard cases'' and refines its discernment. 

This approach enables the student to acquire useful knowledge from the teachers while requiring only limited labeled data.

\subsection{Step 1: Knowledge Acquisition}
\label{sec:knowledge-acquisition}

We define out-of-context news detection knowledge as the reasoning process on the news label. To acquire knowledge from teacher MLLMs, we first prompt them to generate both the predicted label and the corresponding rationale for each news in the training set.

\paragraph{Prompt Design.} From pilot experiment results (Appendix~\ref{sec:pilot-experiments}), we find that teacher MLLMs' zero-shot performance on out-of-context news detection is limited. 
We suspect that this might be due to a lack of news context. 
Inspired by \citet{abdelnabi2022open} and \citet{qi2024sniffer}, we provide the web-retrieved evidence along with the news to mitigate this. The prompt being used is shown as follows.

\begin{tcolorbox}[title=Prompt, colback=gray!10, colframe=darkgray, left=4pt, right=4pt, top=2pt, bottom=2pt]
\setlength{\parskip}{0.1em}
\textcolor{blue}{// Task description}

Does the following \textbf{text description} and the \textbf{attached image} come from the same news post? To help with your judgment, I'll give you the \textbf{visual entities} detected from the attached image and the \textbf{searched captions} of the attached image. 
The searched captions are separated by '<and>'. 
However, if the visual entities or the searched captions is not useful or is empty, 









\end{tcolorbox}

\begin{tcolorbox}[colback=gray!10, colframe=darkgray, left=4pt, right=4pt, top=2pt, bottom=2pt]
just ignore it and give your own prediction.

\textcolor{blue}{// Output formatting}

Please respond with 'yes, the image is rightly used' if there is a semantic match between the text description and the attached image, or 'no' if there are semantic inconsistencies.

Please also give your reason.

\textcolor{blue}{// News input and evidence}

\textbf{Visual entities}: <visual entities>

\textbf{Searched captions}: <searched captions>

\textbf{Text description}: <original caption>

\textbf{Image}: <image>
\end{tcolorbox}

The evidence we use is collected by \citet{abdelnabi2022open}.
Specifically, the searched captions are retrieved by querying the Google Vision API~\citep{google-vision-api} with the image; the visual entities are detected using Google Vision API as well. Although the authors also provide images retrieved by Google Custom Search API using the caption as a query, our experiments show that including them does not improve the zero-shot performance of teacher MLLMs. On the contrary, it slightly degrades the performance and increases the input length. Therefore, we decide to not include them in the final prompt.

For each news in the training set, we obtain the predicted label and rationale from two teacher MLLMs using the above prompt and image. In the next step, the outputs will serve as the ground-truth text that the student MLLM is trained to generate.

\subsection{Step 2: Multi-Teacher Knowledge Distillation}

After obtaining the knowledge from two teacher MLLMs, we aim to fuse and distill their knowledge into a student MLLM. Inspired by curriculum learning~\citep{bengio2009curriculum}, we increase the learning difficulty step by step,
under the assumption that once the student has acquired a solid foundation of knowledge, it can be further refined through learning on more challenging examples. 

To achieve this, based on whether the teachers have reached a consensus on the predicted label, we categorize the news items into two groups. 
We only obtain the \textbf{ground-truth label} for news on which two teacher models have \emph{conflicting} predictions, as we consider these to be "hard cases".
This strategy significantly reduces annotation costs by only annotating the most valuable examples.
Next, we conduct a two-stage knowledge distillation to first learn global knowledge and then cultivate more focused complementary knowledge.

\paragraph{Stage 1: Global Knowledge Learning.} In the first stage, to quickly familiarize the student model with teachers' knowledge, we apply LoRA fine-tuning~\citep{hu2022lora} to the student MLLM on \emph{all} the news items in the training set. 
The input $x = (x_{\mathrm{img}}, x_{\mathrm{prompt}})$ to the student model is the news image and the caption formatted by the prompt in Section~\ref{sec:knowledge-acquisition}:
\begin{equation}
\label{eq:student-input}
x_{\mathrm{prompt}} = \mathrm{Prompt}(x_{\mathrm{cap}}, x_{\mathrm{ent}}, x_{\mathrm{scap}}).
\end{equation}

For each item, the targeted output $y$ is either \one the predicted label plus rationale from one teacher MLLM when both teachers agree, or \two the predicted label plus rationale from the teacher that gives the correct label when the teachers disagree. Note that the correct label is determined by the ground truth label we obtain.
This is illustrated by the two double-ended arrows in Figure~\ref{fig:model}. 

We only train the LoRA adapter (consisting of low rank matrices $A_{i}\in \mathbb{R}^{r\times d}$, $B_{i}\in \mathbb{R}^{d\times r}$), which preserves the original model parameters $\theta$ and reduces the training cost: 
\begin{equation}
\begin{aligned}
\label{eq:lora}
\theta ' = \theta + \Delta \theta, & \quad \Delta \theta = \mathrm{LoRA}(\phi),\\
\phi = & \{A_{i},B_{i}\}.
\end{aligned}
\end{equation}

The optimization objective is the sum of cross-entropy loss at each token's place between the model output $z$ and the targeted output $y$, as described in Eq.~\ref{eq:lora-loss}:
\begin{equation}
\label{eq:lora-loss}
\mathcal{L}_{\mathrm{LoRA}} = - \frac{1}{B}\frac{1}{N}\sum_{i=1}^{B}\sum_{t=1}^{N}\log{P(y_{t}|y_{<t}, x')},
\end{equation}
where $B$ is the batch size and $N$ is the number of tokens. Specifically, $P(y_{t}|y_{<t}, x) = \frac{\exp{(z_{tv})}}{\sum_{k}^{V}\exp{(z_{tk})}}$ where $z$ is the model output, $V$ is the vocabulary size and $v$ is the token id of $y_{t}$. $z_{tk}$ refers to the logit of the $k$-th word for the $t$-th output token.

Through this token-level generated text optimization, the student learns to emulate the teacher's reasoning process for detecting out-of-context news. Although the teacher’s reasoning and predicted labels may sometimes be incorrect - since we do not introduce annotations for news items where teachers agree - we believe this stage lays a foundational understanding of the out-of-context news detection logic for the student.

\paragraph{Stage 2: Complementary Knowledge Fusion.} As previously mentioned, some news items receive \emph{conflicting} predictions from both teacher MLLMs. We assume that these news items are ``hard cases'', where targeted training may boost the student MLLM’s performance in out-of-context news detection. The training data for the second stage only includes the ``hard case''.
This is illustrated by the double-ended arrow in Figure~\ref{fig:model}. The input format, $x = (x_{\mathrm{img}}, x_{\mathrm{prompt}})$, is the same as the previous stage.

To help the student model better reflect the nuances between correct reasoning and incorrect reasoning and better apply it to detect the news, we employ DPO training~\citep{rafailov2023direct} together with LoRA fine-tuning on a second LoRA adapter of parameters $\phi'$, which is formulated in Eq.~\ref{eq:lora2}:
\begin{equation}
\begin{aligned}
\label{eq:lora2}
\theta '' = &\ \theta ' + \Delta \theta', \\
\Delta \theta' = \mathrm{LoRA}&(\phi') + \mathrm{DPO}(\phi'), \\
\phi' = & \{A_{i}',B_{i}'\},
\end{aligned}
\end{equation}
where $\gamma$ and $\alpha$ are the weights to adjust the learning focus. This modular design of having multiple LoRA adapters can separate global knowledge from complementary knowledge, making it easy to plug in or out any new knowledge.

We use obtained ground truth label to determine the correctness of the teacher's output. In LoRA fine-tuning, the targeted output $y$ is the correct teacher's output. In DPO, the student is encouraged to output the preferred answer, and is penalized for outputting the rejected answer. Here, we assign the preferred answer, $y_{w}$, as the correct teacher's output, and the rejected answer, $y_{l}$, as the incorrect teacher's output. 

DPO aims to increase the probability margin of generating the preferred answer and the rejected answer. Since these ``hard cases'' are difficult to predict by the teacher models, we adopt this learning strategy to help the student uncover their patterns and thus become more discerning. 
The optimization objective is the sum of LoRA fine-tuning loss (in Eq.~\ref{eq:lora-loss}) and the DPO training loss (in Eq.~\ref{eq:dpo-loss}):
\begin{equation}
\label{eq:total-loss}
\mathcal{L}_{\mathrm{total}} = \gamma \cdot \mathcal{L}_{\mathrm{LoRA}} + \alpha \cdot \mathcal{L}_{\mathrm{DPO}},
\end{equation}
\begin{align}
\label{eq:dpo-loss}
\mathcal{L}_{\mathrm{DPO}} 
= -\mathbb{E}_{(x, y_{w}, y_{l})\sim B} \Bigg[ 
& \log \sigma \Big( \beta \log \frac{\pi_{\theta ''}(y_{w} | x)}{\pi_{\mathrm{ref}}(y_{w} | x)} \nonumber \\
& \quad - \beta \log \frac{\pi_{\theta ''}(y_{l} | x)}{\pi_{\mathrm{ref}}(y_{l} | x)} \Big) \Bigg],
\end{align}
\begin{equation}
\label{eq:policy-output}
\pi(y|x) = \prod_{t=1}^{T} P(y_{t}|y_{<t}, x),
\end{equation}
where $\pi_{\theta ''}$ denotes the student model that serves as the policy model in this context, and $\pi_{\mathrm{ref}}=\pi_{\theta '}$ denotes the student model from the last stage that serves as a reference model for evaluating relative improvements in this context. Their output is the generation probability of the targeted sequence (assuming that the length is $T$) as described in Eq.~\ref{eq:policy-output}. $\sigma(\cdot)$ denotes the logistic function and $\beta$ is the sensitivity parameter.

Through collaboration, 
LoRA fine-tuning helps the student reinforce correct reasoning, while DPO steers the student away from incorrect reasoning, aiding in the discovery of underlying patterns in challenging cases. Together, these two learning strategies teach the student model to be more discerning.

\section{Experimental Design}

\subsection{Dataset}
\label{sec:dataset}

Following prior works~\cite{qi2024sniffer, lakara2024mad, abdelnabi2022open}, we evaluate our approach on the largest out-of-context news detection benchmark, NewsCLIPpings~\cite{luo-etal-2021-newsclippings}. This dataset is derived from VisualNews~\cite{liu2021visual}, a news image captioning benchmark dataset. The news is collected from four agencies --- The Guardian, BBC, The Washington Post and USA Today. The out-of-context news is generated by replacing the original image by a semantically similar one from another news event. 

\citet{abdelnabi2022open} extends the NewsCLIPpings Merged-Balanced subset with collected textual and visual evidence, including 
\one the detected visual entities, \two text-retrieved images and \three image-retrieved image captions. We use \one and \three in our work. We follow existing works~\cite{qi2024sniffer, lakara2024mad, abdelnabi2022open} and evaluate our approach on the Merged-Balanced subset. It contains 71,072 training samples, 7,024 validation samples and 7,264 test samples.

\subsection{Implementation Details}

\paragraph{Knowledge Acquisition.} The teacher model's rationales are obtained by running Qwen2-VL-72B and InternVL-2.5-78B on 4$\times$ and 2$\times$ NVIDIA Tesla A100 (80GB) GPUs, respectively. The Qwen2-VL-72B inference is accelerated by \texttt{vllm}. The InternVL-2.5-78B inference is accelerated by \texttt{lmdeploy}. This step takes around 70 hours (per model) to complete.

\paragraph{Multi-Teacher Knowledge Distillation.} The experiments are carried out on 1$\times$ NVIDIA Tesla A100 (80GB) GPU. For Stage 1, the batch size is 16. The learning rate is $2e^{-5}$. The warm-up ratio is 0.1. We use LLaVA-v1.6-Vicuna-7B as the student model. The student model is trained for 3 epochs. For Stage 2, the batch size is 8. The learning rate is $5e^{-7}$. The warm-up ratio is 0.1. In Eq.~\ref{eq:total-loss}, the DPO weight $\alpha$ is 0.5 and the LoRA weight $\gamma$ is 0.3.  In Eq.~\ref{eq:dpo-loss}, $\beta$ is set to 0.1. The student model is trained for 1 epoch. 
For both stages, the LoRA rank $r$ is set to 128 and the LoRA alpha is set to 256. Stage 1 takes around 30 hours to complete.
Stage 2 takes around 1 hour to complete. Among all the training data, 8.61\% (6,122 news items) receive conflicting predictions from the teachers. When both teachers' predictions agree, we use Qwen2's output as knowledge because it outperforms InternVL on the test set.

\paragraph{Evaluation.} The experiments are carried out on 1$\times$ NVIDIA Tesla A100 (80GB) GPU. The evaluation takes around 3 hours.

\subsection{Baselines}

As summarized in Section~\ref{sec:rw-ooc}, existing approaches can be divided into 3 categories. Therefore, we select 8 representative methods as baselines from all 3 categories, including methods based on \one \textbf{Pre-trained VLMs:} VisualBert, CLIP, DT-Transformer, SoftLogic; \two \textbf{Auxiliary Information:} CCN, SoR-Stance; and \three \textbf{MLLMs/LLMs:} Chat-OOC, Sniffer. We detail these below.

\textbf{VisualBert}~\cite{li2019visualbert} is a visual-language pre-trained model that implicitly aligns text elements and image regions.
It is used by NewsCLIPpings as one of the base models.

\textbf{CLIP}~\cite{radford2021learning} is a pre-trained visual-language model that uses contrastive learning to learn image and text representations. It is used by NewsCLIPpings as the other base model.

\textbf{DT-Transformer}~\cite{papadopoulos2023synthetic} employs CLIP to encode multimodal inputs and synthetic information, and then applies Transformers to enhance the multimodal features interaction.

\textbf{SoftLogic}~\cite{ma2024interpretable} proposes a logic regularization approach for better detection explainability.

\textbf{CCN}~\cite{abdelnabi2022open} retrieves textual and visual evidence of the image-caption pair for a more thorough consistency check.

\textbf{SoR-Stance}~\cite{yuan-etal-2023-support} extracts stances from evidence and calculates the support-refutation score based on the co-occurrence relations of named entities.

\textbf{Chat-OOC}~\cite{shalabi2024leveraging} fine-tunes a chat-based MLLM, MiniGPT-4, with a cross entropy loss on out-of-context news detection.

\textbf{Sniffer}~\cite{qi2024sniffer} employs two-stage instruction tuning on InstructBLIP~\cite{dai2024instructblip} and uses external knowledge for contextual consistency check.

\section{Result and Analysis}

We evaluate the performance of our proposed \model and analyze the contribution of each of its components. In addition, we present the sensitivity analysis and case study.

\subsection{Overall Performance}

Table~\ref{tab:main-results} shows the performance of \model and the baseline models, where Acc. denotes the accuracy of all the news in the test set. Prec., Rec., and F1 denotes precision, recall and F1-score of \emph{out-of-context news} in the test set.

\begin{table}[t]
    \centering
    \small
    \caption{Performance of \model and baselines on NewsCLIPpings. The best performance is highlighted in \textbf{bold}. The second best performance is \underline{underlined}. $\Delta$ \emph{Improve} represents our method's relative improvements over the second best performance in percentage.}
    \label{tab:main-results}
    \resizebox{\columnwidth}{!}{
    \begin{tabular}{lcccc}
        \toprule
        Method & Acc. & Prec. & Rec. & F1 \\
        \midrule
        \multicolumn{5}{c}{Pre-trained VLMs-based} \\
        \midrule
        VisualBert & 60.23 & 50.39 & 62.73 & 55.90  \\
        CLIP & 58.63 & 38.85 & 54.16 & 45.26  \\
        DT-Transformer & 77.09 & 77.92 & 75.61 & 76.77  \\   
        SoftLogic & 70.51 & - & - & - \\
        \midrule
        \multicolumn{5}{c}{Auxiliary information-based} \\
        \midrule
        CCN & 84.65 & 84.80 & 84.55 & 84.67 \\
        SoR-Stance & 87.10 & 88.60 & 85.92 & 87.27 \\
        \midrule
        \multicolumn{5}{c}{MLLMs/LLMs-based} \\
        \midrule
        Chat-OOC (13B) & 80.00 & 78.27 & 83.18 & 80.67 \\
        Sniffer (13B) & \underline{88.39} & \underline{91.35} & \underline{86.90} & \underline{89.01}  \\
        \model (7B) & \textbf{90.04} & \textbf{91.59} & \textbf{88.19} & \textbf{89.86}   \\	
        \rowcolor{gray!20}
        $\Delta$ \emph{Improve} & 1.87\% & 0.26\% & 1.48\% & 0.95\% \\ 
        \bottomrule
    \end{tabular}
    }
\end{table}

From Table~\ref{tab:main-results}, we see that our \model outperforms the baselines in all four metrics. Notably, it outperforms SOTA by 1.87\% in overall detection accuracy and 1.48\% in recall.
This indicates that our model is more effective than the baselines at detecting out-of-context news, which is especially essential in a real-world deployment. 
Moreover, our model contains only 7B parameters, which is nearly half the size of the second-best baseline. This demonstrates both the effectiveness and efficiency of our approach.


Across different categories of approaches, we observe that methods leveraging MLLMs or LLMs generally outperform others across all evaluation metrics. We believe that MLLMs/LLMs have great potential of detecting out-of-context news with effective instruction tuning. In contrast, earlier methods based on pre-trained VLMs yield inferior performance. We attribute this to the fact that news images and captions often contain fine-grained semantics that are not well captured by generic VLMs. Moreover, given the nature of news content, incorporating external news evidence or auxiliary information clearly benefits the detection - an aspect that these earlier methods fail to utilize.

In summary, \model outperforms the baselines in all metrics with notable efficiency in model size.

\subsection{Ablation Study}
\label{sec:ablation-study}

To demonstrate the effectiveness of different components in \model, we present an ablation study in this section. Specifically, we ablate \textbf{Step 1 + Step 2}, \textbf{Step 2}, \textbf{DPO} in Step 2 and \textbf{LoRA FT} in Step 2. The results are listed in Table~\ref{tab:ablation-study}.

\begin{table}[t]
    \centering
    \small
    \caption{Evaluation results on ablating \textbf{Step 1 + Step 2}, \textbf{Step 2}, \textbf{DPO (Step 2)} and \textbf{LoRA FT (Step 2)}. The best performance is highlighted in \textbf{bold}. The second best performance is \underline{underlined}.}
    \label{tab:ablation-study}
    \begin{tabular}{p{2.8cm}cccc}
        \toprule
        Method & Acc. & Prec. & Rec. & F1 \\
        \midrule
        w/o Step 1 + Step 2 & 76.53 & 76.48 & 76.56 & 76.63 \\
        w/o Step 2 & 89.66 & 90.67 & \underline{88.41} & 89.53 \\
        w/o DPO (Step 2) & \underline{89.91} & 90.72 & \textbf{88.91} & \underline{89.80} \\
        w/o LoRA FT (Step 2) & 89.55 & \textbf{93.84} & 84.65 & 89.01 \\
        \model & \textbf{90.04} & \underline{91.59} & 88.19 & \textbf{89.96} \\
        \bottomrule
    \end{tabular}
\end{table}

\paragraph{Does the framework improve the student's performance?} 
We observe that removing Step 1 and Step 2 results in the most significant performance degradation, which corresponds to the zero-shot performance of the student model. This indicates that the teacher models’ predictions and rationales provide valuable knowledge for the student to learn from, and that LoRA fine-tuning is effective in facilitating this knowledge transfer. The 13.49\% improvement in detection accuracy achieved by \model over w/o Step 1 + Step 2 highlights the effectiveness of our approach to improve the student MLLM's detection performance. 

\paragraph{Does multi-teacher knowledge fusion further enhance the learning?} 
Furthermore, removing Step 2 results in a slight performance drop. Similarly, removing DPO also leads to a minor decrease in accuracy, suggesting that DPO is able to improve the student model on detecting harder cases.
We assume that the teacher models are sufficiently capable of providing accurate knowledge for most news items, which allows the student model to learn adequate knowledge from Step 1 (Global Knowledge Learning). However, given the volume of news in the real world, removing either component will impact the detection of a substantial amount of out-of-context news items.
Lastly, we observe that removing LoRA fine-tuning from Step 2 results in inferior performance compared to not introducing Step 2. We conjecture that applying DPO in isolation may confuse the student model regarding generation targets. However, with proper guidance (LoRA fine-tuning), DPO can realize its full potential and benefit learning on more difficult examples.

In summary, the ablation study confirms the importance of both knowledge acquisition and multi-teacher knowledge distillation, as well as DPO and LoRA fine-tuning in the second step. 

\subsection{Sensitivity Analysis}

This section gives an in-depth analysis on how the hyperparameters LoRA rank $r$ (Step 1 and Step 2), DPO weight $\alpha$ and sensitivity parameter $\beta$ affect the performance of \model.

\begin{figure}[htbp]
  \centering
  \subfloat[LoRA rank $r$ (Step 1)]{\includegraphics[width=0.5\linewidth]{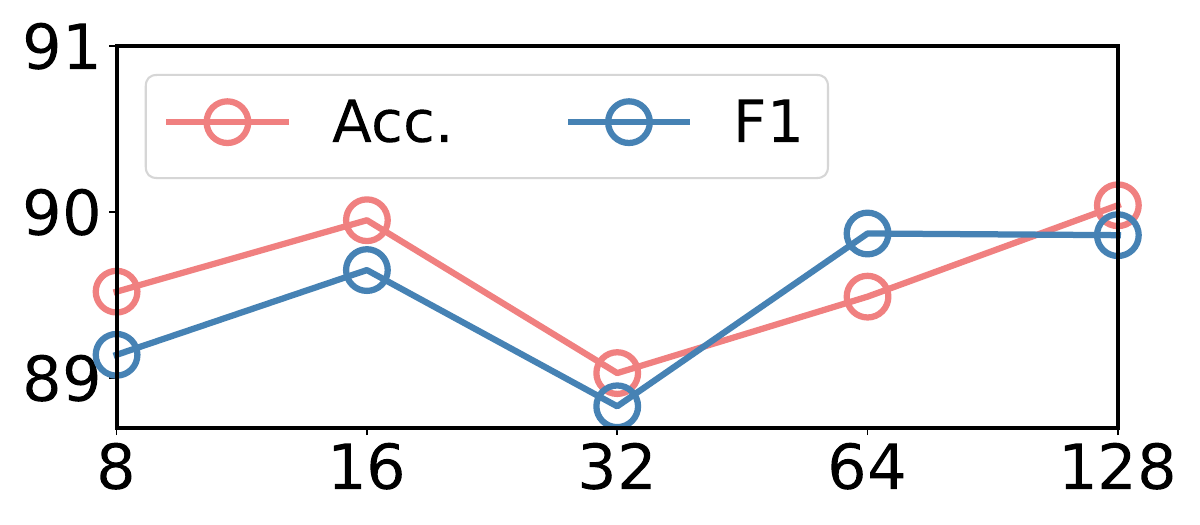}}
  \subfloat[LoRA rank $r$ (Step 2)]{\includegraphics[width=0.5\linewidth]{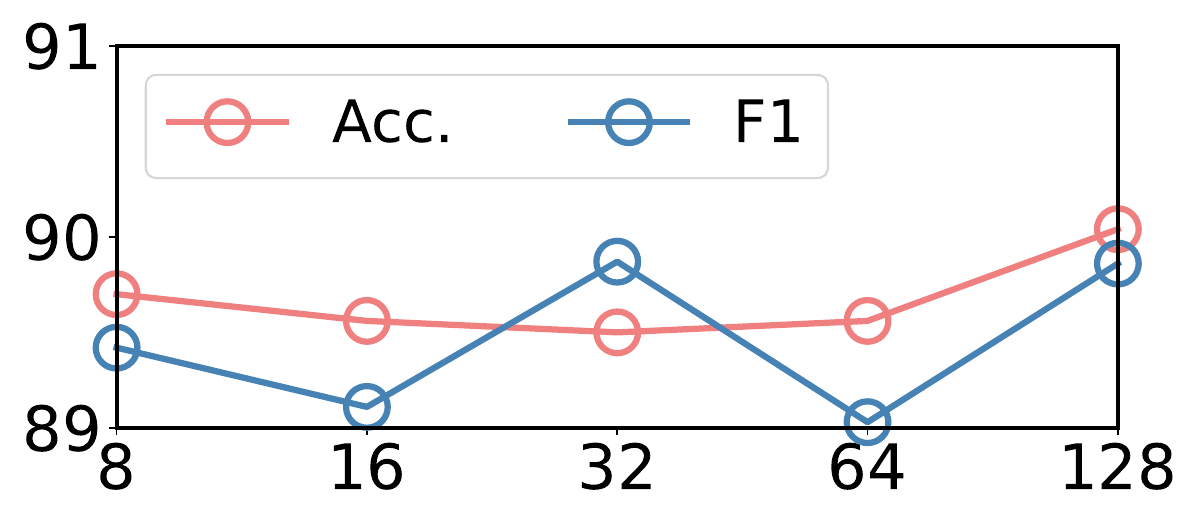}}
  \hfill
  \subfloat[Sensitivity parameter $\beta$]{\includegraphics[width=0.5\linewidth]{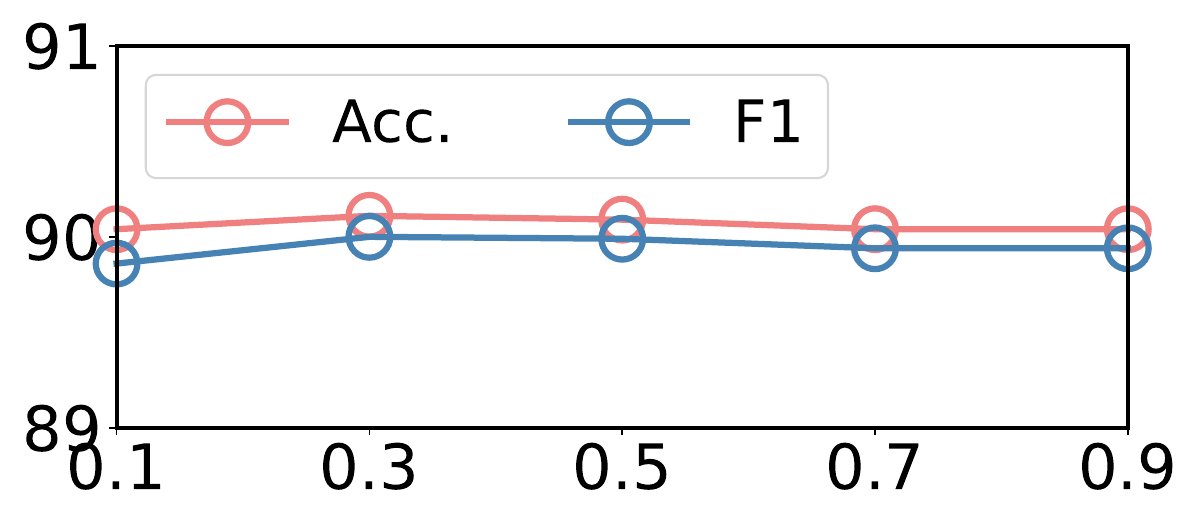}}
  \subfloat[DPO weight $\alpha$]{\includegraphics[width=0.5\linewidth]{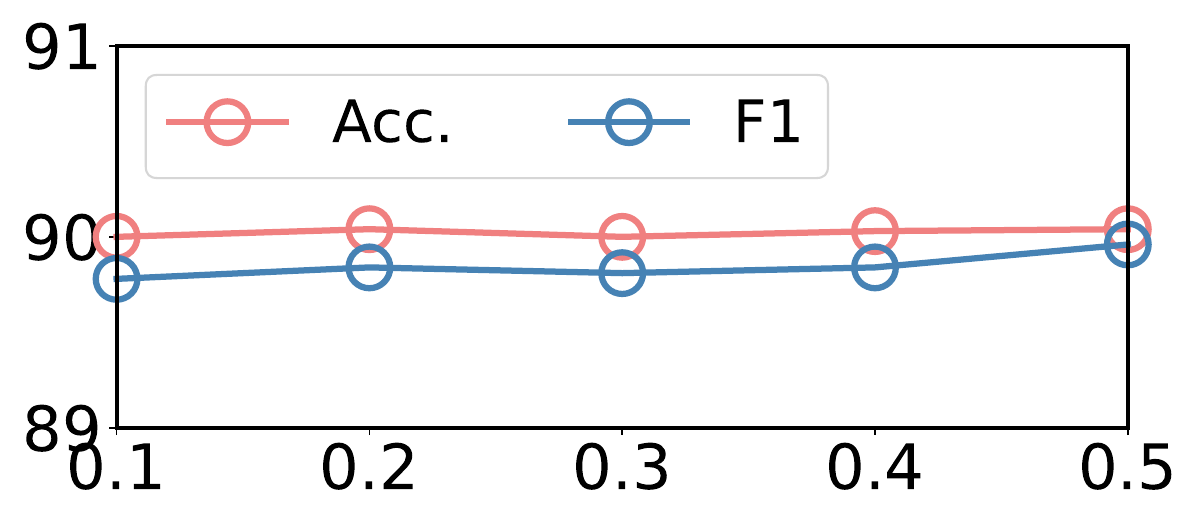}}
  \caption{\model's performance with different LoRA rank $r$, DPO weight $\alpha$ and sensitivity parameter $\beta$ values.}
  \label{fig:sensitivity}
\end{figure}

We observe that changing the LoRA rank $r$ in Step 1 and Step 2 affects model performance differently. In Step 1, when $r$ is 32, both the overall accuracy and F1 reach the lowest; while both larger and smaller values lead to improved performance. In Step 2, when $r$ is 32, the accuracy reaches the lowest while F1 reaches the highest. The highest performance is observed at $r=$128. Regardless of the variance, the performances are very close. We conjecture that the student model is less sensitive to the LoRA rank within a reasonable range.

We find that different values of $\beta$ have little impact on the student model's overall performance. We conjecture that the log-probability difference (in Eq.~\ref{eq:dpo-loss}) is already substantial, resulting in close outputs after the outer logistic transformation. Therefore, the predicted label is not strongly affected by $\beta$. Moreover, different DPO weight $\alpha$ has limited impact on the student model's accuracy; but when it is 0.5, F1 reaches the highest. We believe that the higher weight of DPO loss will emphasize the preference learning, which may better help the model learn from hard cases.



\subsection{Case Study}

In this section, we present a case study (examples in Appendix~\ref{sec:case-study}) to help understand how our framework improves the student model's reasoning. 

In Figure~\ref{fig:case-1}, we observe that without applying \model, the student model recognizes the text on the banner in the image but only associates it with a general protest or demonstration. As a result, it incorrectly assumes that the news image and caption refer to the same event. However, the student model after applying \model focuses on the mention of the ``steel industry'' on the banner, prompting it to examine the details more carefully and correctly conclude that the news image is out-of-context.

In Figure~\ref{fig:case-2}, the student without applying \model extracts wrong text from the image, and subsequently assumes no semantic match between the news caption and the news image. In contrast, after applying \model, the student model correctly extracts the relevant text from the image, which is crucial for the accurate detection.

\section{Conclusion}

In this paper, we propose a novel multi-MLLM knowledge distillation framework, \model, for out-of-context news detection. \model first prompts multiple teacher MLLMs to generate label predictions and corresponding rationales for the news and then incorporate two-stage knowledge distillation to convey knowledge to student MLLMs. Experimental results validate the effectiveness and efficiency of our approach. These advantages enhance \model's applicability across a wider deployment spectrum.


\section*{Limitations}
Despite the effectiveness of our proposed \model, our work has several limitations. First, we do not explore a broader range of teacher MLLMs, because Qwen2-VL-72B and InternVL-2.5-78B are currently the only available models that satisfy our performance requirements for teacher models. As more high-performing MLLMs become available, we plan to incorporate them into our framework. Additionally, we evaluate our approach on the one largest dataset for this task. In future work, we intend to collect real-world news data to further assess the generalizability and effectiveness of our approach.

\bibliography{custom}

\appendix

\section{Appendix}
\label{sec:appendix}

\subsection{Pilot Experiments}
\label{sec:pilot-experiments}

In Table~\ref{tab:pilot-experiments}, we show that the teacher MLLMs (take Qwen2-VL-72B as an example) performance is affected by the prompt.

\begin{table*}[!ht]
    \centering
    \small
    \caption{Qwen2-VL-72B's performance on the test set.}
    \label{tab:pilot-experiments}
    \begin{tabular}{p{12cm}cc}
        \toprule
        Prompt & Acc. & F1 \\
        \midrule
        Does the following text description and the attached image come from the same news post? Please respond with 'yes' if there is a semantic match and 'no' if there are semantic inconsistencies. Text description: <original\_caption> & 82.85 & 82.88 \\
        \midrule
        Does the following text description and the attached image come from the same news post? I'll give you the visual entities detected from the attached image. But if the visual entities is not useful or is None, just ignore it and give your own prediction.
        
        Please respond with 'yes, the image is rightly used' if there is a semantic match between the text description and the attached image, or 'no' if there are semantic inconsistencies.
        
        Please also give your reason.
        
        Visual entities: <visual\_entities>
        
        Text description: <original\_caption> & 87.26 & 86.63 \\
        \midrule
        Does the following text description and the attached image come from the same news post? To help with your judgment, I'll give you the searched captions of the attached image. The searched captions are separated by '<and>'. However, if the searched captions is not useful or is empty, just ignore it and give your own prediction.
        
        Please respond with 'yes, the image is rightly used' if there is a semantic match between the text description and the attached image, or 'no' if there are semantic inconsistencies.
        
        Please also give your reason.
        
        Searched captions: <inv\_search\_captions>
        
        Text description: <original\_caption> & 90.80 & 90.50 \\
        \midrule
        Does the following text description and the attached image come from the same news post? To help with your judgment, I'll give you the visual entities detected from the attached image and the searched captions of the attached image. The searched captions are separated by '<and>'. However, if the visual entities or the searched captions is not useful or is empty, just ignore it and give your own prediction.
        
        Please respond with 'yes, the image is rightly used' if there is a semantic match between the text description and the attached image, or 'no' if there are semantic inconsistencies.
        
        Please also give your reason.
        
        Visual entities: <visual\_entities>
        
        Searched captions: <inv\_search\_captions>
        
        Text description: <original\_caption> & 91.11 & 90.81 \\
        \midrule
        Does the following text description and the attached image come from the same news post? To help with your judgment, I'll give you the visual entities detected from the attached image, the searched captions of the attached image, and the searched images of the text description. The searched captions are separated by '<and>'. The searched images are presented in the format of text and are separated by '<and>'. However, if the visual entities or the searched captions or the searched images is not useful or is empty, just ignore it and give your own prediction.
        
        Please respond with 'yes, the image is rightly used' if there is a semantic match between the text description and the attached image, or 'no' if there are semantic inconsistencies.
        
        Please also give your reason.
        
        Visual entities: <visual\_entities>
        
        Searched captions: <inv\_search\_captions>
        
        Searched images: <dir\_search\_images>
        
        Text description: <original\_caption> & 65.31 & 71.19 \\
        \bottomrule
    \end{tabular}
\end{table*}

\subsection{Case Study Examples}
\label{sec:case-study}

\begin{figure*}[t]
  \centering
  \includegraphics[width=\textwidth]{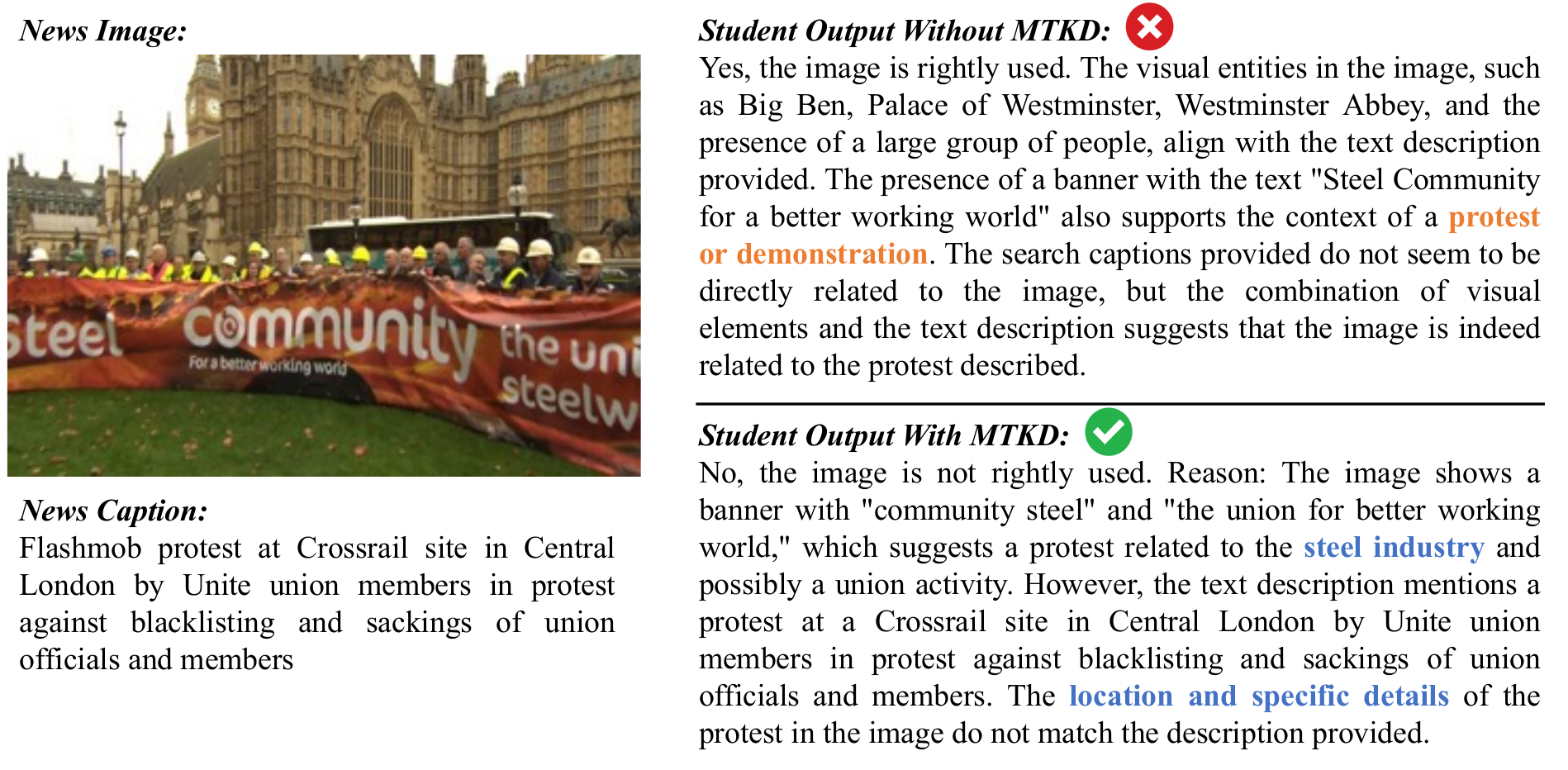}
  \caption{Case 1 of student model's output without and with \model.}
  \label{fig:case-1}
\end{figure*}

\begin{figure*}[t]
  \centering
  \includegraphics[width=\textwidth]{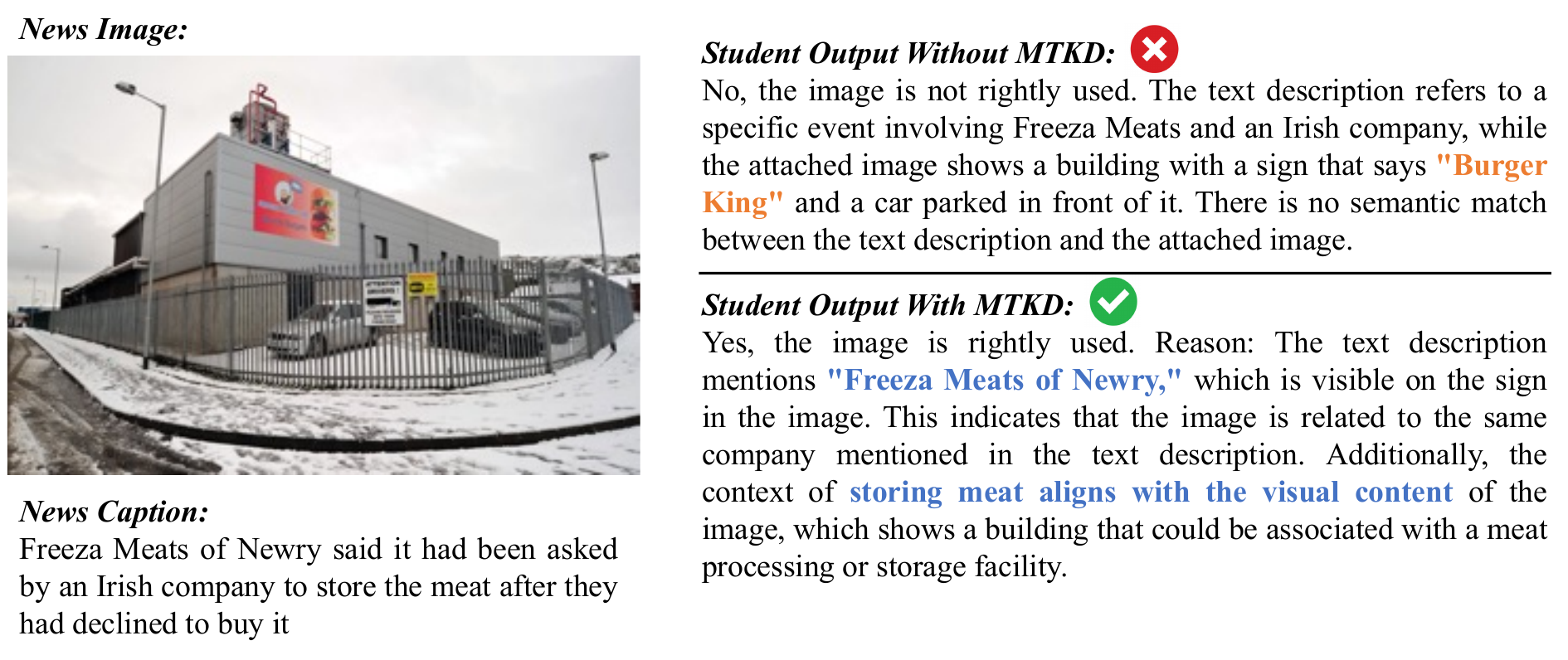}
  \caption{Case 2 of student model's output without and with \model.}
  \label{fig:case-2}
\end{figure*}

\end{document}